\documentclass[journal]{IEEEtran}

\usepackage{cite}
\usepackage{amsmath,amssymb,amsfonts}
\usepackage{booktabs}
\usepackage{array}
\usepackage{url}
\usepackage{graphicx}
\usepackage{tikz}
\usetikzlibrary{positioning,arrows.meta,fit,calc}
\usepackage[hidelinks]{hyperref}

\newif\ifieeesubmitted
\ieeesubmittedfalse
\newcommand{\btid}{\texttt{bt\_id}}
\newcommand{\traceid}{\texttt{trace\_id}}

\begin{document}

\title{The Two-Ledger Model: Transaction-Grain Cost Accounting for Multi-Agent AI Systems Across Clouds, Models, and Systems of Record}

\author{Vivek~Acharya%
\thanks{V. Acharya is an Independent Researcher, Dallas, TX, USA (e-mail: vacharya@bu.edu; ORCID: 0009-0002-0860-9462).}%
\thanks{Manuscript prepared November 2026.}}

\markboth{Preprint, November 2026}%
{Acharya: The Two-Ledger Model}

\maketitle

\begin{abstract}
Enterprises run business transactions through multi-agent AI systems spanning clouds, large language models (LLMs), orchestration harnesses, and systems of record. Finance receives AI cost only at settlement grain, aggregated by SKU and charge period. The business question --- \emph{what did this order, claim, or case cost in AI?} --- cannot be answered from billing data, because providers never see enterprise transaction boundaries. The FinOps Open Cost and Usage Specification (FOCUS) gives billing data one schema across providers but remains settlement-grain even as it adds AI token and requester attribution. We formalize this grain mismatch, derive seven requirements, and propose the \emph{Two-Ledger Model}: a settlement ledger of FOCUS-conformant billing data joined to an attribution ledger of invocation-grain usage records built on OpenTelemetry GenAI telemetry. A layered \emph{correlation spine} (trace context, A2A/MCP metadata, business-key correlation, probabilistic matching) assigns each record to its business transaction with recorded method and confidence; a versioned rate card prices usage in near real time; period reconciliation ties both ledgers to the invoice, and the unexplained residual identifies shadow AI. Costs are reported per order, claim, or outcome with agentic cost metrics. A simulation with known ground truth shows the spine sustaining cost coverage above 99\% as trace propagation falls to zero, the reconciliation residual recovering planted unmetered spend within 0.24 percentage points while flagging provider billing errors, and mid-flight gating cutting p99 transaction cost by up to 61\% at a 2--4\% false-pause rate. We propose the attribution ledger as an invocation-grain FOCUS profile for standardization.
\end{abstract}


\IEEEpeerreviewmaketitle

\section{Introduction}
\IEEEPARstart{A}{n} enterprise order-to-cash (O2C) transaction no longer lives in one system. A customer email lands in a customer relationship management (CRM) system, where an intake agent extracts the order intent. A pricing agent checks availability and discounts in SAP. An order-creation agent writes the sales order into Oracle Fusion Cloud. An exception agent opens and works a case in ServiceNow when a credit hold fires. Each agent may call a different frontier large language model (LLM) --- one hosted on AWS Bedrock, one on Azure OpenAI, one on Google Vertex AI --- through a different orchestration harness, with retries, tool calls, and sub-agent delegation along the way. The business sees one transaction: order \texttt{ORD-48812}. The chief financial officer asks one question: \emph{what did that transaction cost in AI?}

Today no one can answer it. Finance receives AI spend at \emph{settlement grain}: monthly invoices and daily billing exports aggregated by provider, stock-keeping unit (SKU), and charge period. Engineering holds \emph{runtime grain}: traces and logs that record each model call but carry no price and no tie to the invoice. The transaction --- the unit in which the business actually accounts for its work --- exists in neither dataset. Scale this across every multi-cloud, multi-LLM, multi-harness workload in a large enterprise, and AI becomes a growing line item that cannot be attributed to the work it performed.

The standards landscape has moved quickly on the settlement side. FOCUS\footnote{FOCUS is an open specification published under the Creative Commons Attribution 4.0 (CC-BY-4.0) license by the FinOps Open Cost and Usage Specification project, a Linux Foundation project series \cite{focus-license}; Linux Foundation trademark usage rules apply.}, the FinOps Open Cost and Usage Specification developed under the FinOps Foundation (a Linux Foundation project), gives billing data one schema across cloud, SaaS, and AI providers; version 1.4 was ratified in June 2026, and version 1.5, targeted for December 2026, is scoped to add AI model and token cost visibility, a price sheet dataset, and cost attribution to people and agents \cite{focus-spec,aispendops15}. On the runtime side, the OpenTelemetry GenAI semantic conventions standardize how a model call is described in telemetry, including input, output, cached, and reasoning token counts \cite{otel-genai}. On the agent side, the Model Context Protocol (MCP) \cite{mcp} and the Agent2Agent (A2A) protocol, contributed by Google to the Linux Foundation in June 2025 \cite{a2a-lf,a2a-spec}, standardize how agents reach tools and each other.

Each of these standards is necessary. None of them, alone or together, yields transaction-grain cost. Billing data lacks the transaction; telemetry lacks the price; agent protocols lack both. The gap between them is where AI cost attribution falls today, and it is the gap this paper addresses.

We ask three research questions:

\begin{itemize}
\item \textbf{RQ1.} Why can settlement-grain billing standards such as FOCUS not answer transaction-grain cost questions for agentic workloads, and what exactly is missing?
\item \textbf{RQ2.} What minimal, standards-aligned architecture gives every business transaction an auditable, reconcilable AI cost, across clouds, models, harnesses, and systems of record?
\item \textbf{RQ3.} How should that cost be expressed in the language in which businesses already track their work --- orders, claims, cases, hires --- and tied back, period by period, to the invoices finance actually pays?
\end{itemize}

Our contributions are:

\begin{enumerate}
\item \textbf{A formalization of the grain mismatch} between settlement-grain billing data and transaction-grain business accounting, with seven requirements (R1--R7) that any solution must meet (Section~\ref{sec:problem}).
\item \textbf{The Two-Ledger Model}, a reference architecture that joins a FOCUS-conformant settlement ledger to an invocation-grain attribution ledger built on OpenTelemetry GenAI telemetry, with an enterprise LLM gateway as the primary metering point, a versioned rate card for near-real-time pricing, and a component-level deployment reference architecture (Section~\ref{sec:tlm}).
\item \textbf{The correlation spine}, a layered correlation strategy --- trace context, A2A/MCP metadata propagation, business-key correlation sets, and probabilistic matching --- that assembles one business transaction from spans scattered across systems of record that cannot all propagate trace context (Section~\ref{sec:spine}).
\item \textbf{Business-language cost semantics}: a mapping from invocation-grain cost to a process taxonomy such as the APQC Process Classification Framework, and a set of agentic cost metrics --- effective cost per transaction, tail (p95) cost per outcome, token efficiency, coordination overhead, and rework cost --- together with a period reconciliation discipline whose variance doubles as a shadow-AI signal (Sections~\ref{sec:language} and \ref{sec:recon}).
\item \textbf{An analytical evaluation, a synthetic O2C walkthrough, a simulation study, and a realization analysis} demonstrating the model end to end --- the simulation measuring correlation coverage and accuracy under degraded instrumentation, reconciliation recovery of planted unmetered spend and provider billing errors, and the tail-cost effect of mid-flight gating, all on synthetic data with known ground truth --- including three sourcing models (assemble, accelerate, or productize) and an illustrative best-of-breed market mapping of the components, and a discussion of alignment with the FOCUS~1.5 AI scope, including the proposal that the attribution ledger itself be published as a FOCUS-shaped, invocation-grain dataset using the specification's extension (\texttt{x\_}) columns (Sections~\ref{sec:scenario}--\ref{sec:discussion}).
\end{enumerate}

The remainder of the paper is organized as follows. Section~\ref{sec:background} reviews FOCUS, runtime telemetry standards, agent protocols, and related work. Section~\ref{sec:problem} formalizes the problem. Sections~\ref{sec:tlm}--\ref{sec:recon} present the Two-Ledger Model. Section~\ref{sec:scenario} walks through a synthetic O2C scenario. Section~\ref{sec:eval} evaluates the model analytically, and Section~\ref{sec:sim} reports a simulation study on synthetic data with known ground truth. Section~\ref{sec:realization} examines how the architecture is realized --- assembled by the enterprise, delivered as an accelerator, or productized --- against a best-of-breed market exemplar. Section~\ref{sec:discussion} discusses governance integration, limitations, and standardization outlook. Section~\ref{sec:conclusion} concludes.

\section{Background and Related Work}
\label{sec:background}

\subsection{FOCUS: Normalizing the Bill}
FOCUS defines one schema --- one set of column definitions and one vocabulary --- for cost and usage data across technology providers \cite{focus-spec}. Table~\ref{tab:focus-versions} summarizes its release trajectory. Version 1.0 (June 2024) established the core columns and brought the four leading cloud providers to a common export format. Version 1.1 (November 2024) deepened multi-cloud granularity. Version 1.2 (May 2025) extended the specification to software-as-a-service (SaaS) and platform-as-a-service (PaaS) billing, which is what lets a direct model-vendor invoice sit in the same table as a hyperscaler export. Version 1.3 (December 2025) added allocation columns for splitting shared-resource cost, a contract-commitment dataset, and dataset timestamping with completeness flags \cite{sdtimes13}. Version 1.4 (June 2026), the current ratified version, added invoice reconciliation, billing-period finality, and commitment detail; it contains no AI-specific columns \cite{focus-spec,aispendops15}.

\begin{table}[!t]
\caption{FOCUS release trajectory \cite{focus-spec,sdtimes13,aispendops15}.}
\label{tab:focus-versions}
\centering
\begin{tabular}{@{}lll@{}}
\toprule
\textbf{Version} & \textbf{Ratified / target} & \textbf{Headline additions} \\
\midrule
1.0 & Jun.\ 2024 & Core schema; four major clouds \\
1.1 & Nov.\ 7, 2024 & Granular multi-cloud columns \\
1.2 & May 29, 2025 & SaaS and PaaS billing support \\
1.3 & Dec.\ 4, 2025 & Allocation columns; contract \\
    &               & commitments; data finality flags \\
1.4 & Jun.\ 4, 2026 & Invoice reconciliation; commitment \\
    &               & detail; no AI-specific columns \\
1.5 & Dec.\ 2026 (target) & AI model/token visibility; price \\
    &               & sheet; person/agent attribution \\
\bottomrule
\end{tabular}
\end{table}

Two properties of FOCUS matter for this paper. First, the schema can already carry token-denominated AI charges using existing columns: token counts in \texttt{ConsumedQuantity} with \texttt{ConsumedUnit}, the priced model offering in \texttt{SkuId} and \texttt{SkuPriceId}, the token type in \texttt{SkuMeter}, the service classified under the \emph{AI and Machine Learning} service category, and reserved model throughput in the commitment-discount columns \cite{focus-ai,glassity14}. The gap before 1.5 is agreed usage of these columns, not missing schema. Second, the specification reserves an \texttt{x\_} prefix for custom columns, which guarantees that enterprise extensions cannot collide with future standard columns \cite{glassity14}. We use both properties in Section~\ref{sec:tlm}.

The 1.5 scope, announced ahead of ratification, adds requester-attribution columns --- reported in preview analyses as \texttt{PrincipalId}, \texttt{CredentialId}, and \texttt{RequesterDetails} --- so that billing data can name the person, workload, or agent behind a charge without a join to observability data \cite{tokeneconomics15,aispendops15}. This is, to our knowledge, the first time a billing standard explicitly acknowledges non-human requesters. We return to why identity attribution, while necessary, still does not close the transaction-grain gap (Section~\ref{sec:problem}).

\subsection{Runtime Telemetry: OpenTelemetry GenAI and Trace Context}
The OpenTelemetry (OTel) GenAI semantic conventions define a vendor-neutral vocabulary for model telemetry: span names and attributes for model operations (\texttt{gen\_ai.\allowbreak provider.\allowbreak name}, \texttt{gen\_ai.\allowbreak operation.\allowbreak name}, requested and response model identifiers) and token accounting (\texttt{gen\_ai.\allowbreak usage.\allowbreak input\_tokens}, \texttt{gen\_ai.\allowbreak usage.\allowbreak output\_tokens}, with cached and reasoning tokens where the model reports them), plus metrics for token usage and operation duration \cite{otel-genai}. The conventions now extend to agent and workflow execution and to MCP operations, and are maintained in a dedicated repository with an overall status of \emph{Development} \cite{otel-genai}. Trace assembly across process and service boundaries rests on W3C Trace Context, which standardizes the \texttt{traceparent} header for context propagation \cite{w3c-trace}.

\subsection{Agentic Systems and Agent Protocols}
LLM-based agents interleave reasoning with tool use \cite{react} and increasingly operate as conversing multi-agent ensembles \cite{autogen}. Two open protocols shape the enterprise substrate: MCP standardizes how a model-backed application reaches tools and data sources \cite{mcp}, and A2A, an open protocol created by Google and now a Linux Foundation project with a v1.0.0 specification, standardizes secure communication between opaque agents across frameworks and vendors \cite{a2a-lf,a2a-spec}. For cost accounting, these protocols matter in a specific way: they are the seams across which correlation context must survive. An agent step that crosses an A2A boundary or an MCP tool call that fans out to another model must carry enough metadata that its spend can still be assigned to the originating business transaction. A recent IETF Internet-Draft proposes signature-chained delegation tokens for AI agents that carry per-chain spend ceilings, explicitly not running balances, with spend enforcement out of band and cost recording and billing outside its scope \cite{aip}; Section~\ref{sec:ext} relates our attribution tokens to that design.

\subsection{Business Process Context}
Enterprises track work in process language. Cross-system flows such as order-to-cash, procure-to-pay, and hire-to-retire are typically anchored in a process taxonomy; the APQC Process Classification Framework (PCF) is the most widely used cross-industry example \cite{apqc}. Process platforms have long solved a structurally similar correlation problem: BPMN~2.0 defines correlation keys that bind asynchronous messages to the right process instance using business data (an order number, a case number) when no technical context travels with the message \cite{bpmn}. We adapt this device directly in Section~\ref{sec:spine}.

\subsection{Related Practitioner Work and Positioning}
Practitioner work has started on parts of the problem. Microsoft's FinOps guidance demonstrates normalizing Azure OpenAI billing into FOCUS and computing token unit economics from billing data \cite{ms-aoai}, and proposes meter maps that enrich opaque FOCUS rows with AI attributes (token type, product) via a join against a curated SKU dictionary \cite{ms-tokens}. Vendor tooling has shown that token usage can be written as FOCUS~1.4 rows today using existing columns \cite{glassity14}. All of this operates within or against the settlement ledger; none of it reaches the transaction, because the transaction is not present in billing data to begin with. Our contribution is to treat that absence as structural, not incidental, and to design the second ledger and the join that the question requires. To our knowledge, no prior work formalizes transaction-grain AI cost attribution for multi-agent systems spanning multiple systems of record, nor connects the FOCUS, OpenTelemetry GenAI, MCP, and A2A standards into one accounting architecture.

\section{Problem Formalization}
\label{sec:problem}

\subsection{Definitions}
\begin{itemize}
\item \textbf{Business transaction (BT).} A unit of work the business accounts for as one thing: an order, a claim, a support case, a hire. Each BT $b$ carries a stable identifier \btid{} and a type $\tau(b)$ drawn from the enterprise process taxonomy (e.g., a PCF element). A BT spans systems of record and time; \texttt{ORD-48812} is a BT.
\item \textbf{Outcome.} The completed result of a BT against which the business measures value (order fulfilled, claim adjudicated, case resolved). Cost per outcome is the unit economics the business ultimately wants.
\item \textbf{Agent step.} One agent's bounded contribution to a BT: a span (or span subtree) executed by an identified agent within a harness, possibly delegated across an A2A boundary.
\item \textbf{Model invocation.} A single call $i$ to a model endpoint, with usage vector $u_i = (u_{i,c})_{c \in C}$ over token classes $C$ (input, output, cache read, cache write, reasoning), a priced offering $s_i$ (provider, model, SKU), a timestamp $t_i$, an outcome status (success, retried, failed), and any per-call tool fees $f_i$ (e.g., metered search calls).
\item \textbf{Settlement record.} A row of billing data --- in this paper, a FOCUS row --- aggregating charges by provider, SKU, charge period, and billing dimensions such as sub-account and tags.
\end{itemize}

\subsection{The Grain Mismatch}
Let $\mathcal{I}$ be the set of model invocations in a period and $\mathcal{F}$ the set of settlement records for the same period. Billing generation applies an aggregation $\sigma: \mathcal{I} \rightarrow \mathcal{F}$ that is many-to-one and keyed on provider-visible dimensions only. The assignment of invocations to business transactions is a correlation $\beta: \mathcal{I} \rightarrow \mathcal{B}$, where $\mathcal{B}$ is the set of BTs. The business question is the cost of $I(b) = \{\, i : \beta(i) = b \,\}$.

The structural observation is simple: $\beta$ is not recoverable from $\mathcal{F}$. The provider never sees \btid; the harness, the retry, the sub-agent delegation, and the downstream system of record are all invisible to billing generation, so $\sigma$ destroys exactly the information $\beta$ needs. Tags do not fix this: a tag identifies a workload or a team, not a transaction instance, and tags cannot in general vary per request on token-metered services. Identity attribution (FOCUS 1.5's requester columns) narrows the aggregate --- it says \emph{which agent} spent --- but an agent serves thousands of transactions per period, so identity grain is still not transaction grain. Conversely, runtime telemetry recovers $\beta$ naturally --- the invocation happens inside the transaction's execution --- but telemetry carries no settlement truth: no negotiated rate, no commitment amortization, no invoice.

So the answer requires two datasets with one join, not a better version of either dataset alone. This answers RQ1, and it motivates the requirements below.

\subsection{Requirements}
\begin{table}[!t]
\caption{Requirements for transaction-grain AI cost accounting.}
\label{tab:reqs}
\centering
\begin{tabular}{@{}p{0.6cm}p{7.2cm}@{}}
\toprule
\textbf{ID} & \textbf{Requirement} \\
\midrule
R1 & \textbf{Provider neutrality.} One accounting model across clouds, model vendors, SaaS AI, and self-hosted inference. \\
R2 & \textbf{Transaction grain.} Every BT has a cost object; every attributable invocation maps to exactly one BT, across systems of record. \\
R3 & \textbf{Business language.} Costs roll up to the enterprise process taxonomy and to outcome units (per order, per claim, per case). \\
R4 & \textbf{Agentic completeness.} Fan-out, delegation (A2A), tool calls (MCP), retries, and failed attempts are all captured and classifiable. \\
R5 & \textbf{Economic fidelity.} Negotiated rates, commitment amortization (provisioned throughput), cache and batch tiers, per-call tool fees, and shared-infrastructure allocation are reflected. \\
R6 & \textbf{Reconcilability.} Attributed cost ties to invoices period by period; the variance is measured and decomposable. \\
R7 & \textbf{Timeliness.} A near-real-time cost estimate exists at transaction close; settlement truth corrects it at period close. \\
\bottomrule
\end{tabular}
\end{table}

Table~\ref{tab:reqs} states requirements R1--R7. Four failure modes stress them: \emph{bypass} (calls that never pass a metering point --- shadow AI); \emph{broken propagation} (context lost at a SaaS or protocol boundary); \emph{shared pools} (provisioned throughput, graphics processing unit (GPU) clusters, vector databases serving many BTs); and \emph{settlement lag} (invoices arriving weeks after the work). The architecture must degrade gracefully under each.

\section{The Two-Ledger Model}
\label{sec:tlm}

\begin{figure*}[!t]
\centering
\begin{tikzpicture}[
  font=\small,
  box/.style={draw, rounded corners=1pt, align=center, inner sep=4pt, minimum height=7mm},
  wide/.style={box, minimum width=14.4cm},
  arr/.style={-{Latex[length=2.2mm]}, thick}
]
\node[wide] (agents) {Multi-agent workloads: harnesses, agents, MCP tools, A2A peers \emph{(across CRM, SAP, Oracle Fusion, ServiceNow, custom apps)}};
\node[box, below=7mm of agents, xshift=-36mm, minimum width=5.6cm] (gw) {Enterprise LLM gateway\\(primary metering point)};
\node[box, right=14mm of gw, minimum width=4.2cm] (prov) {Model providers\\(Bedrock, Azure OpenAI, Vertex, direct)};
\node[box, below=7mm of gw, minimum width=5.6cm] (attr) {\textbf{Attribution ledger} (invocation grain)\\OTel GenAI fields; \btid, \traceid,\\agent id, token classes, est.\ cost};
\node[box, below=7mm of prov, minimum width=4.2cm] (focus) {\textbf{Settlement ledger}\\FOCUS-conformant exports\\+ vendor invoices};
\node[box, left=8mm of attr, minimum width=1.8cm] (rate) {Rate card\\(versioned)};
\node[wide, below=7mm of attr.south -| agents] (join) {Join \& allocation layer: correlation spine $\beta$ \;$\cdot$\; shared-cost allocation \;$\cdot$\; period reconciliation $\Delta$};
\node[wide, below=7mm of join] (biz) {Business semantics: process taxonomy (PCF) roll-up \;$\cdot$\; cost per transaction / per outcome \;$\cdot$\; $c_{95}$, TE, CO, RW \;$\cdot$\; showback and budget gates};
\draw[arr] (agents.south -| gw.north) -- (gw.north);
\draw[arr] (gw) -- (prov);
\draw[arr] (gw) -- (attr);
\draw[arr] (prov) -- (focus);
\draw[arr] (rate) -- (attr);
\draw[arr] (attr.south) -- (attr.south |- join.north);
\draw[arr] (focus.south) -- (focus.south |- join.north);
\draw[arr] (join) -- (biz);
\end{tikzpicture}
\caption{The Two-Ledger Model. Every model call is metered at the gateway into an invocation-grain attribution ledger priced by a versioned rate card; FOCUS-conformant billing data forms the settlement ledger; a join and allocation layer correlates, allocates, and reconciles; a business semantics layer expresses cost in process language.}
\label{fig:arch}
\end{figure*}
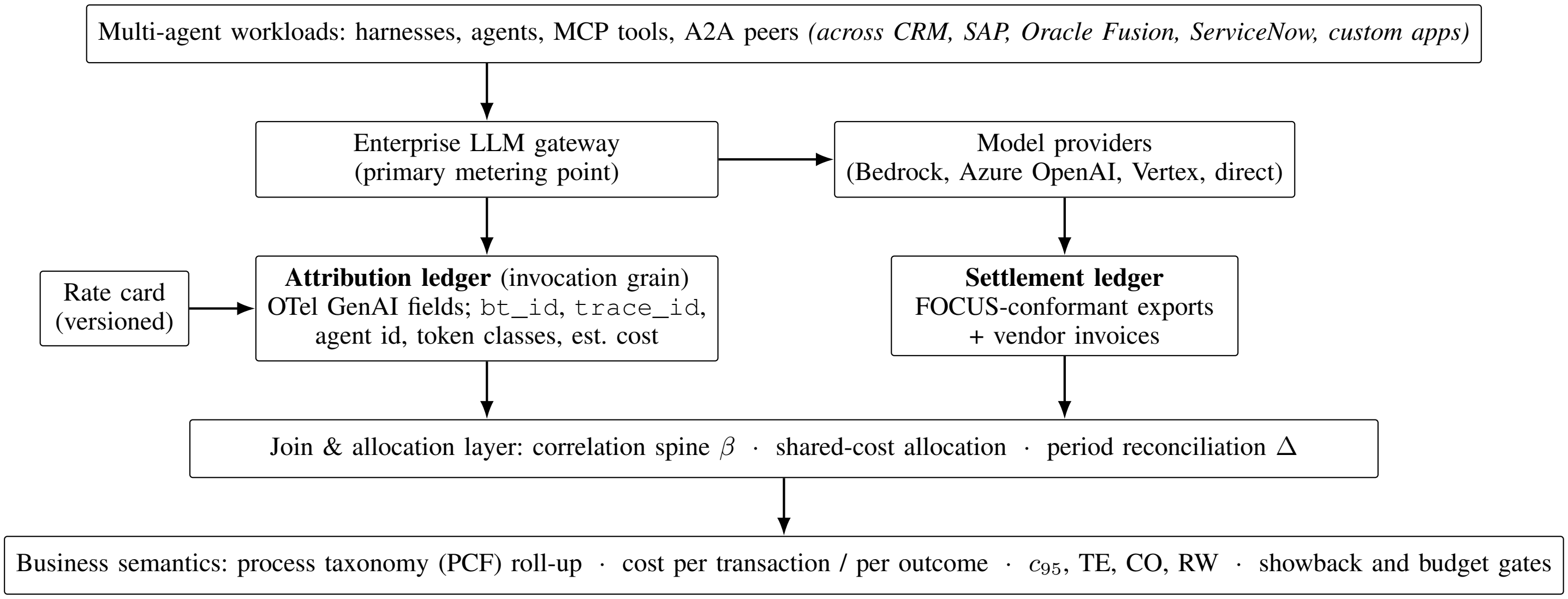

The Two-Ledger Model answers RQ2. The organizing idea comes from accounting. A general ledger cannot answer sub-ledger questions; no company runs accounts receivable out of the general ledger. FOCUS is the general ledger of technology spend. Transaction-grain AI cost is a sub-ledger problem, and the sub-ledger must be built where the transactions are visible: at runtime. Figure~\ref{fig:arch} shows the architecture; we describe each component.

\subsection{Settlement Ledger}
The settlement ledger is a FOCUS-conformant lake of all technology billing data: hyperscaler exports, marketplace charges, and --- via FOCUS 1.2's SaaS support --- direct model-vendor invoices normalized into the same columns \cite{focus-spec,focus-ai}. Commitment purchases (provisioned model throughput, reserved GPU capacity) enter through the commitment-discount columns and, from 1.3, the contract-commitment dataset; 1.3's timestamps and completeness flags gate downstream processing on data finality \cite{sdtimes13}. Nothing in this layer is novel. Its job is completeness: this is the ledger that ties to cash.

\subsection{Attribution Ledger}
The attribution ledger records one row per model invocation (and one per metered tool call), written at runtime. Table~\ref{tab:schema} gives the core schema and its mapping to OpenTelemetry GenAI attributes and FOCUS columns. Three design choices matter.

\begin{table}[!t]
\caption{Attribution-ledger core schema with standards mapping.}
\label{tab:schema}
\centering\footnotesize
\setlength{\tabcolsep}{3pt}
\begin{tabular}{@{}p{1.8cm}p{3.0cm}p{2.9cm}@{}}
\toprule
\textbf{Field} & \textbf{Runtime source (OTel GenAI / W3C)} & \textbf{Settlement analog (FOCUS)} \\
\midrule
\traceid, span & \texttt{traceparent} \cite{w3c-trace} & --- (absent) \\
\btid, bt type & propagated baggage / correlation spine (\S\ref{sec:spine}) & \texttt{x\_bt\_id}, \texttt{x\_bt\_type} \\
outcome id & business event & \texttt{x\_outcome\_id} \\
tenant, app, env & resource attributes & \texttt{Tags} \\
agent id, harness & \texttt{gen\_ai.*} agent spans & 1.5 \texttt{PrincipalId} / \texttt{RequesterDetails} \\
credential & workload identity & 1.5 \texttt{CredentialId} \\
provider, model & \texttt{gen\_ai.provider.\allowbreak name}, model attrs & \texttt{SkuId}, \texttt{SkuPriceId}, SKU price details \\
token counts by class & \texttt{gen\_ai.usage.*} & \texttt{ConsumedQuantity} + \texttt{ConsumedUnit}; type in \texttt{SkuMeter} \\
tool-call fees & span events & per-call SKU rows \\
status / retry flag & span status & --- \\
est.\ list / eff.\ cost & rate card ($\S$\ref{sec:cost}) & \texttt{ListCost} / \texttt{EffectiveCost} \\
rate card version & gateway & \texttt{x\_rate\_card\_ver} \\
correlation method, confidence & spine level L1--L4 & \texttt{x\_corr\_method}, \texttt{x\_corr\_conf} \\
\bottomrule
\end{tabular}
\end{table}

First, \textbf{the gateway is the primary metering point}. Routing every LLM call --- all providers, all harnesses --- through an enterprise gateway makes the attribution ledger a side effect of normal operation rather than a per-team instrumentation project, and makes bypass an exception to detect rather than a norm to tolerate. Where a gateway is impossible (embedded SaaS copilots, provider-managed agents), an SDK/sidecar writes the same records, and provider usage application programming interfaces (APIs) backfill what neither can see; those rows carry a lower-confidence source marker.

Second, \textbf{the schema is the OpenTelemetry GenAI vocabulary plus business keys}. Token classes, model identity, and operation timing come straight from the conventions \cite{otel-genai}; the additions are \btid, the transaction type, the outcome identifier, and the correlation metadata of Section~\ref{sec:spine}. This keeps the ledger implementable by configuration on any OTel-instrumented stack.

Third, \textbf{the attribution ledger is itself published FOCUS-shaped}. Because FOCUS already has columns that carry token usage and reserves the \texttt{x\_} prefix for extensions \cite{focus-ai,glassity14}, the gateway can emit the attribution ledger as invocation-grain FOCUS rows with \texttt{x\_bt\_id}, \texttt{x\_trace\_id}, and companions. One query engine and one column vocabulary then serve both grains. This design choice is also a standardization proposal, which we take up in Section~\ref{sec:discussion}. A privacy boundary is built into the schema: the ledger carries usage metadata --- token counts, identities, timings, status --- and never prompt or response content. Content has no role in cost accounting and does not enter this path.

\subsection{Rate Card Service}
\label{sec:ratecard}
The rate card is a versioned internal price book: list prices per token class per SKU, negotiated rates, cache-read and batch tiers, context-window price tiers where providers use them, per-call tool fees, and amortization rules for committed capacity. Every attribution row is priced on write with the then-current version, giving R7's near-real-time estimate; reconciliation (Section~\ref{sec:recon}) corrects rate drift at period close. A price sheet dataset is in the announced FOCUS 1.5 scope \cite{aispendops15}; when ratified, it becomes the rate card's settlement-side anchor.

\subsection{A Deployment Reference Architecture}
\label{sec:deploy}

\begin{figure*}[!t]
\centering
\begin{tikzpicture}[
  font=\scriptsize,
  comp/.style={draw, rounded corners=1pt, align=center, inner sep=4pt},
  Lc/.style={comp, text width=5.0cm, anchor=west},
  Mc/.style={comp, text width=4.1cm, anchor=west},
  Rc/.style={comp, text width=4.6cm, anchor=west},
  arr/.style={-{Latex[length=2mm]}, thick},
  barr/.style={arr, dashed}
]
\node[Lc] (ex) at (0,0) {\textbf{Execution plane}\\orchestration harnesses \& agents · MCP tools · A2A peers --- every call carries \texttt{traceparent} + \btid{} baggage};
\node[Lc] (gw) at (0,-2.2) {\textbf{Metering plane}\\LLM gateway cluster: workload authN · budget gate · token capture · price-on-write\\\emph{backfill:} SDK/sidecar spans, provider usage-API pollers (lower confidence)};
\node[Lc] (al) at (0,-4.5) {\textbf{Attribution ledger}\\invocation-grain stream $\rightarrow$ table, FOCUS-shaped with \texttt{x\_} business keys};
\node[Lc] (ca) at (0,-6.4) {\textbf{Correlation \& allocation engines}\\spine $\beta$ (L1--L4, confidence) · shared-pool shares $\alpha_p$};
\node[Lc] (ug) at (0,-8.2) {\textbf{Unit economics \& gates}\\$\hat{c}$, $c_{95}$, TE/CO/RW marts \& APIs · budget-gate decisions};
\node[Mc] (sor) at (6.1,0) {\textbf{Systems of record}\\CRM · SAP · Oracle Fusion · \mbox{ServiceNow} --- business events with keys};
\node[Mc] (rc) at (6.1,-2.2) {\textbf{Rate card service}\\versioned: list · negotiated · commitment rules · tiers};
\node[Mc] (reg) at (6.1,-4.5) {\textbf{Taxonomy \& identity registry}\\PCF codes · tag schema · agent \texttt{PrincipalId}s};
\node[Rc, dashed] (ext) at (11.4,0) {\textbf{Cloud \& AI providers} (external)\\inference endpoints · billing generation};
\node[Rc] (sl) at (11.4,-2.2) {\textbf{Settlement lake}\\FOCUS exports + vendor invoices, gated on finality flags};
\node[Rc] (recon) at (11.4,-6.4) {\textbf{Reconciliation engine}\\$\Delta$ decomposition · unmetered (shadow-AI) report};
\node[Rc] (close) at (11.4,-8.2) {\textbf{Close \& showback}\\invoice tie-out pack · chargeback · working-group export};
\draw[arr] (ex.south) -- (gw.north);
\draw[arr] (gw.south) -- (al.north);
\draw[arr] (al.south) -- (ca.north);
\draw[arr] (ca.south) -- (ug.north);
\draw[arr] (ext.south) -- (sl.north);
\draw[arr] (sl.south) -- (recon.north);
\draw[arr] (recon.south) -- (close.north);
\draw[arr] (ca.east) -- (recon.west);
\draw[arr] (rc.west) -- (rc.west -| gw.east);
\draw[arr] (sor.west) -- ++(-0.5,0) |- ([yshift=3mm]ca.east);
\draw[arr] (reg.west) -- ++(-0.25,0) |- ([yshift=-4mm]gw.east);
\draw[barr] (ug.west) -- ++(-0.45,0) |- (gw.west);
\end{tikzpicture}
\caption{Deployment reference architecture. Left: the runtime attribution path (execution $\rightarrow$ gateway metering $\rightarrow$ attribution ledger $\rightarrow$ correlation/allocation $\rightarrow$ unit economics). Right: the settlement path (provider billing $\rightarrow$ FOCUS lake $\rightarrow$ reconciliation $\rightarrow$ close). Middle: shared services --- business events from systems of record, the versioned rate card, and the taxonomy and identity registry. The dashed return edge is the budget-gate feedback from cost SLOs to the gateway; inference traffic from the gateway to provider endpoints is omitted for clarity (see Fig.~\ref{fig:arch}).}
\label{fig:deploy}
\end{figure*}
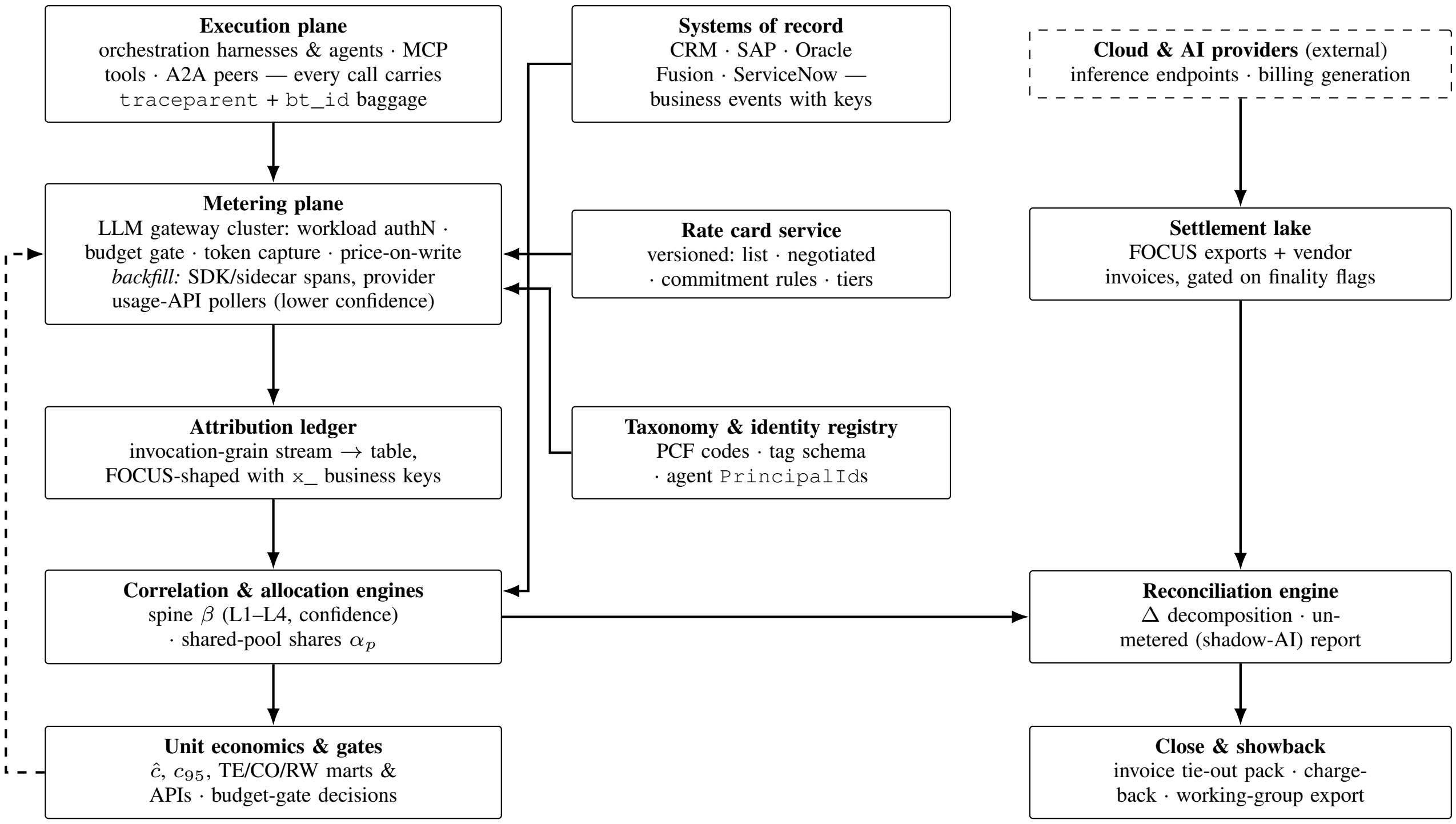

Two questions decide what an enterprise actually builds here. \emph{Is this a system or a gateway feature?} It is a reference architecture: cooperating planes with specified records at every seam, of which the gateway is the primary metering point --- necessary, but one component among eight (Table~\ref{tab:components}). \emph{Is it a product?} No single one. Every component maps to a category most enterprises already run --- an LLM gateway, OpenTelemetry collectors, a lakehouse, a FinOps platform, an identity provider, a policy engine --- so the proposal is compositional. The contribution is the contracts between the components: which identifiers and which records cross each edge of Fig.~\ref{fig:deploy}, so that independently sourced parts add up to one auditable accounting system.

\begin{table*}[!t]
\caption{Reference-architecture components: responsibilities, interfaces, and typical realizations.}
\label{tab:components}
\centering\footnotesize
\setlength{\tabcolsep}{4pt}
\begin{tabular}{@{}p{2.4cm}p{5.6cm}p{4.3cm}p{3.6cm}@{}}
\toprule
\textbf{Component} & \textbf{Responsibility} & \textbf{Key interfaces} & \textbf{Typical realization} \\
\midrule
1\; LLM gateway & Meters every sanctioned model call; enforces workload authN, policy, and budget gates; stamps agent and credential identity; prices rows on write & In: model calls with trace context; out: attribution rows, gate verdicts; onward: provider endpoints & LLM-gateway--class proxy (Envoy-based or commercial) \\
2\; SDK/sidecar emitters \& usage pollers & Cover paths no proxy sees (embedded copilots, provider-managed agents, batch); backfilled rows carry lower-confidence source markers & OpenTelemetry GenAI spans; provider usage APIs & OTel SDK and Collector; scheduled collectors \\
3\; Attribution ledger & Invocation-grain system of record for usage; FOCUS-shaped rows with \texttt{x\_} business keys & Append stream in; SQL out to engines and marts & Lakehouse table (Iceberg/Delta class) fed by a stream \\
4\; Settlement lake & All billing truth: FOCUS exports and vendor invoices, gated on data-finality flags & Provider FOCUS exports (1.2 SaaS path for model vendors); invoice feeds & FinOps platform store or a lakehouse zone \\
5\; Rate card service & Versioned price book: list and negotiated rates, commitment-amortization rules, cache/batch/context tiers, tool fees & Pricing API to the gateway; settled-rate restatement to reconciliation; FOCUS 1.5 price-sheet sync when ratified & Internal service over a config repository \\
6\; Taxonomy \& identity registry & One schema for every join key: PCF process codes, the tag taxonomy, agent \texttt{PrincipalId} and credential registry & Configuration to gateway, engines, and tagging pipelines & Workload identity provider + metadata catalog \\
7\; Correlation, allocation, reconciliation engines & $\beta$ at L1--L4 with recorded confidence; $\alpha_p$ for shared pools; period $\Delta$ decomposition and the unmetered report & Both ledgers in; unit-economics marts and the close pack out & Stream and batch jobs on the lakehouse \\
8\; Unit-economics \& gate layer & $\hat{c}$, $c_{95}$, TE/CO/RW per process type; showback and chargeback; turns cost-SLO breaches into gateway gate decisions & BI and APIs out; policy push back to the gateway & FinOps/BI semantic layer + a policy engine \\
\bottomrule
\end{tabular}
\end{table*}

End to end, a transaction moves through the architecture as follows. (1)~A call leaves a harness carrying \texttt{traceparent} and \btid{} baggage; the gateway authenticates the workload identity, checks the budget gate, and forwards to the provider endpoint. (2)~The gateway writes an attribution row priced by the current rate-card version. (3)~Paths the gateway cannot see arrive as SDK spans or usage-API backfill, marked lower-confidence. (4)~Systems of record emit business events --- order created, case opened --- carrying their keys. (5)~The correlation engine resolves $\beta$ through the spine and stamps method and confidence on every row. (6)~The allocation engine applies $\alpha_p$ from shared-pool meters. (7)~Providers deliver FOCUS exports into the settlement lake, gated on finality. (8)~At close, reconciliation computes $\Delta$, restates the period against settled rates, and names the unmetered residual. (9)~Marts publish unit economics, and $c_{95}$ breaches return to the gateway as gate decisions --- the loop that connects accounting back to the Govern and Gate stages of the agentic lifecycle (Section~\ref{sec:discussion}).

Three non-goals set the scope. The architecture stores no prompt or response content; it does not replace an observability platform or a FinOps platform; and it does not standardize agent behavior. It standardizes the accounting records those systems exchange. Section~\ref{sec:realization} returns to this architecture with the market's actual parts.

\subsection{Extensions: Verifiable Lineage, Two-Sided Reconciliation, Mid-Flight Gating}
\label{sec:ext}

Three extensions strengthen the model where finance and runtime control meet. They are described here and, where measurable, evaluated in Section~\ref{sec:sim}.

\textbf{Verifiable lineage (level L0).} The gateway or orchestrator mints, at transaction start, a signed \emph{attribution token} binding the business transaction identifier, type, tenant, issuance time, expiry, and a nonce, signed with a key in the client's key-management service. The token travels in trace baggage and in A2A task metadata or MCP call context. At each delegation, the delegating party mints a child token hash-linked to its parent, so a signature chain runs from the originating transaction to every downstream invocation, including across organizational boundaries. A usage record carrying a verified token is assigned at a level L0 that takes precedence over L1--L4, with the level recorded on the record like any other; an auditor can check a transaction's cost lineage from public keys and the hash chain without trusting the ledger operator. The token differs from capability tokens with static spend ceilings \cite{aip} in three ways: it is bound to the business transaction, it participates in the correlation spine with its assignment recorded alongside confidence stamping, and L0-assigned spend is reported as a distinct, verifiable stratum in reconciliation. In the budget-envelope variant, the token also carries a \emph{running} remaining allowance and delegation depth; a delegate decrements the allowance locally from consumed amounts it returns, and a child token's sub-envelope never exceeds its parent's remainder, so a cost cap propagates with delegated work without a call back to the gateway.

\textbf{Two-sided reconciliation.} With double-entry semantics, each priced usage record posts a debit to its cost object and a credit to a provider-payable accrual keyed by provider, SKU, and period; ingested settlement records clear accruals. The reconciliation of Section~\ref{sec:recon} then runs in both directions: billed consumption with no accrual is the unmetered residual, and accruals left uncleared after finality --- or billed quantities priced inconsistently with the settled price sheet --- form a provider billing-discrepancy register. The same join detects unsanctioned internal consumption and provider-side billing errors.

\textbf{Mid-flight gating.} The budget gate of Section~\ref{sec:language} need not wait for outcomes. A running cost accumulator is kept per open transaction; a forecaster estimates the conditional distribution of completion cost given transaction type, current stage, and accumulated spend, from historical trajectories of the same type; when the forecast probability of breaching the type's cost objective exceeds a threshold $\theta$, the gate acts before further spend --- pausing for approval, capping delegation depth or retries, substituting a cheaper priced offering, or terminating a runaway loop. The trade-off, a paused transaction that would have finished within budget, is measurable and is measured below.

\section{The Correlation Spine}
\label{sec:spine}

Requirement R2 depends entirely on $\beta$: assembling one BT from spans scattered across systems that differ in what context they can carry. No single mechanism survives every boundary in an O2C flow, so we define a layered spine; every attribution row records which layer produced its assignment and with what confidence (Table~\ref{tab:spine}).

\begin{table}[!t]
\caption{The correlation spine.}
\label{tab:spine}
\centering
\begin{tabular}{@{}p{0.5cm}p{2.5cm}p{4.3cm}@{}}
\toprule
\textbf{Lvl} & \textbf{Mechanism} & \textbf{Where it works / fails} \\
\midrule
L1 & W3C \texttt{traceparent} + baggage carrying \btid{} \cite{w3c-trace} & In-process and service-to-service within instrumented estates; fails at opaque SaaS boundaries. \\
L2 & Protocol metadata: \btid{} and trace context carried in A2A task metadata and MCP call context \cite{a2a-spec,mcp} & Across agent and tool seams between cooperating parties; fails when a peer strips metadata. \\
L3 & Business-key correlation sets (BPMN-style): order no., case no., invoice no.\ + system event time windows \cite{bpmn} & Across systems of record (SAP, Oracle Fusion, ServiceNow, CRM) that expose business keys but no trace context; deterministic when keys are unique. \\
L4 & Probabilistic matching: temporal proximity, agent identity, tenant, workload fingerprint; emits confidence $\in [0,1]$ & Last resort for legacy or black-box paths; rows below a confidence floor fall to a pooled cost object, never silently to a BT. \\
\bottomrule
\end{tabular}
\end{table}

Two disciplines make the spine auditable. \emph{Explicit degradation}: an invocation that cannot be assigned at L1--L3 above a confidence threshold is pooled by workload, not forced onto a transaction; forced assignment would corrupt the unit economics it exists to serve. \emph{Confidence-weighted reporting}: transaction cost can be reported with its correlation confidence, and process-level roll-ups can be stratified by spine level, so finance can see how much of a number is deterministic and how much is inferred. Layer L3 deserves a note. It is forty years of integration practice, not something new: BPMN correlation keys applied to cost. What is new is only the requirement that the spine's provenance travel with every dollar.

\section{Cost Semantics}
\label{sec:cost}

\subsection{Transaction Cost}
Let $I(b)$ be the invocations correlated to BT $b$, $u_{i,c}$ the usage of token class $c$ in invocation $i$, $r_{\mathrm{eff}}(s,c,t)$ the effective rate for SKU $s$, class $c$ at time $t$ from the rate card, $f_i$ per-call tool fees, $P_{\mathrm{sh}}$ the shared resources (vector stores, provisioned pools, GPU clusters), $C_p$ the period cost of shared resource $p$, and $\alpha_p(b)$ its allocation share to $b$ (metered where possible --- queries served, tokens drawn from the pool --- policy-based otherwise, using FOCUS 1.3 allocation semantics on the settlement side \cite{sdtimes13}). The estimated effective cost of a transaction is

\begin{equation}
\hat{c}(b) \;=\; \sum_{i \in I(b)} \Big( \sum_{c \in C} u_{i,c}\, r_{\mathrm{eff}}(s_i, c, t_i) \;+\; f_i \Big) \;+\; \sum_{p \in P_{\mathrm{sh}}} \alpha_p(b)\, C_p .
\label{eq:cost}
\end{equation}

For committed capacity (provisioned throughput treated like a reserved instance), the effective rate amortizes the commitment over consumption measured in the commitment's own meter $m_i$ --- provisioned-throughput units or the provider's normalized consumption units, not raw token counts, since token classes are not equivalent units:

\begin{equation}
r_{\mathrm{eff}}^{(p)} \;=\; \frac{C_p^{\mathrm{commit}}}{\sum_{i \in \mathcal{I}_p} m_i}\,,
\label{eq:amort}
\end{equation}

where $\mathcal{I}_p$ is the set of invocations served from commitment $p$ in the period and $m_i = \mu_p(u_i)$ is invocation $i$'s consumption in that commitment's own meter, a provider-defined normalization of the usage vector. For such invocations, the inner term of Eq.~\ref{eq:cost} evaluates as $m_i\, r_{\mathrm{eff}}^{(p)}$. Idle capacity is either held centrally or spread per policy --- a choice that must be explicit, because it decides whether a quiet month makes every transaction look expensive. Marginal cost (list or negotiated per-token rate only) and fully loaded cost (Eq.~\ref{eq:cost}) are both reported; conflating them is a common unit-economics error in practice.

\subsection{Agentic Metrics}
Agentic execution makes cost per outcome a \emph{distribution}, not a number: fan-out, deliberation, retries, and tool loops vary per instance. For a transaction type $\tau$, we therefore manage the tail,

\begin{equation}
c_{95}(\tau) \;=\; Q_{0.95}\big(\{\hat{c}(b) : \tau(b)=\tau\}\big),
\label{eq:p95}
\end{equation}

as a cost service-level objective (SLO) alongside the median. Three decompositions show where money goes inside a transaction:

\begin{align}
\mathrm{TE}(b) &= \frac{u^{\mathrm{final}}_{\mathrm{out}}(b)}{\sum_{i \in I(b)} \sum_{c} u_{i,c}} \quad \text{(token efficiency)} \label{eq:te}\\
\mathrm{CO}(b) &= \frac{\text{cost of coordination spans}}{\hat{c}(b)} \quad \text{(coordination overhead)} \label{eq:co}\\
\mathrm{RW}(b) &= \frac{\text{cost of failed/retried spans}}{\hat{c}(b)} \quad \text{(rework cost)} \label{eq:rw}
\end{align}

where $u^{\mathrm{final}}_{\mathrm{out}}$ counts output tokens of the accepted final result (the denominator deliberately includes input tokens, so values well below one are expected and are comparable within a transaction type, not across types), coordination spans are inter-agent planning, routing, and consensus traffic (as opposed to task-productive spans), and the definitional choices --- what counts as ``final,'' what counts as ``coordination'' --- are fixed per transaction type and versioned like the rate card. TE extends single-call token-efficiency thinking to the whole transaction; CO makes the price of multi-agent deliberation itself a managed line \cite{scf}; RW is the agentic analog of cost of quality.

\section{Accounting in the Language of the Business}
\label{sec:language}

RQ3 asks for cost in business language, not engineering vocabulary. Three bindings do that.

\textbf{Process taxonomy binding.} Each BT type $\tau$ is mapped to an element of the enterprise process taxonomy --- for cross-industry use, the APQC PCF \cite{apqc}; an O2C transaction's steps map onto the corresponding PCF elements across the market-and-sell, deliver, and manage-financial-resources categories. AI cost then rolls up the same hierarchy the business already uses for cycle time and quality: invocation $\rightarrow$ agent step $\rightarrow$ transaction $\rightarrow$ process $\rightarrow$ value stream. The PCF code lives on the attribution row (\texttt{x\_bt\_type}), so the roll-up is a \texttt{GROUP BY}, not a re-mapping project.

\textbf{Outcome binding.} Unit economics are stated per outcome unit the operating business recognizes: cost per order shipped, per claim adjudicated, per case resolved, per candidate screened --- with distributional form (median, p95) rather than averages alone, per Eq.~\ref{eq:p95}.

\textbf{Financial binding.} Showback and chargeback dimensions (cost center, product, tenant) come from the same tag taxonomy applied in both ledgers, so the number the process owner sees and the number finance allocates reconcile by construction. The governing rule is simple: \emph{the join keys are a governance decision, not a tooling decision} --- one taxonomy (tenant, product, process, environment), mandated in traces, tags, and gateway metadata alike.

\section{Reconciliation: Tying the Ledgers Together}
\label{sec:recon}

At period close, for each provider $g$, SKU $s$, and charge period $P$, the settlement ledger's billed amount is compared to the sum of attributed effective cost:

\begin{equation}
\Delta_{g,s,P} \;=\; \mathrm{Billed}_{g,s,P} \;-\; \sum_{i \in \mathcal{I}_{g,s,P}} \hat{c}_i\,, 
\qquad V = \frac{\Delta}{\mathrm{Billed}}\,.
\label{eq:recon}
\end{equation}

Here $\hat{c}_i$ is invocation $i$'s priced cost, the inner term of Eq.~\ref{eq:cost}; for commitment SKUs, the attributed side is the amortized consumption of Eq.~\ref{eq:amort} plus allocated idle. FOCUS 1.4's invoice-reconciliation support and 1.3's finality flags gate the computation on complete data \cite{focus-spec,sdtimes13}. The variance $\Delta$ is then decomposed into: (i) rate drift (rate card vs.\ invoiced rates --- corrected by restating the period's rows against settled rates); (ii) timing (usage near period boundaries); (iii) shared-pool idle (Eq.~\ref{eq:amort} policy effects); and (iv) \textbf{unmetered spend} --- billed consumption with no attribution rows at all. Component (iv) is the shadow-AI report: it names, by provider and SKU, exactly the AI spend occurring outside sanctioned paths. The reconciliation variance ratio $V$ is therefore both an accounting control and a governance control, and it belongs on the same executive page as the unit costs. The operating discipline mirrors financial close: attribution coverage (share of billed AI spend carrying a \btid{} or an explicit pool) trends toward a threshold, and exceptions are worked like unreconciled cash.

\section{Illustrative Scenario: One O2C Transaction}
\label{sec:scenario}

\emph{All figures in this section are synthetic and illustrative. They are constructed to exercise the model; they are not measurements from any production system.}

Order \texttt{ORD-48812} (BT type: O2C) arrives as a customer email. Table~\ref{tab:o2c} lists the attribution-ledger rows the transaction generates across four systems of record, three clouds, and four priced model SKUs (three frontier, one mid-tier), including one retried step. Rates are illustrative blended effective rates per million tokens (Mtok) after commitment amortization.

\begin{table*}[!t]
\caption{Synthetic attribution-ledger rows for BT \texttt{ORD-48812} (illustrative rates; all data synthetic).}
\label{tab:o2c}
\centering\footnotesize
\setlength{\tabcolsep}{4pt}
\begin{tabular}{@{}llllrrrlr@{}}
\toprule
\textbf{Step} & \textbf{System of record} & \textbf{Agent} & \textbf{Provider / model} & \textbf{In tok} & \textbf{Out tok} & \textbf{\$/Mtok (in; out)} & \textbf{Status / corr.} & \textbf{Est.\ cost} \\
\midrule
1 & CRM & intake-agent & Azure OpenAI / frontier & 6{,}200 & 1{,}100 & 2.00; 8.00 & ok / L1 & \$0.0212 \\
2 & SAP & pricing-agent & AWS Bedrock / frontier & 9{,}800 & 1{,}900 & 3.00; 15.00 & ok / L2 (A2A) & \$0.0579 \\
3a & Oracle Fusion & order-agent & Vertex AI / frontier & 7{,}500 & 1{,}400 & 1.25; 10.00 & failed / L3 (order no.) & \$0.0234 \\
3b & Oracle Fusion & order-agent (retry) & Vertex AI / frontier & 7{,}600 & 1{,}500 & 1.25; 10.00 & ok / L3 (order no.) & \$0.0245 \\
4 & ServiceNow & exception-agent & Azure OpenAI / frontier & 12{,}400 & 2{,}600 & 2.00; 8.00 & ok / L3 (case no.) & \$0.0456 \\
5 & (coordination) & supervisor & AWS Bedrock / mid-tier & 5{,}000 & 800 & 0.80; 3.20 & ok / L1 & \$0.0066 \\
--- & --- & tool: entitlement check & metered API & --- & --- & \$0.004 / call $\times$ 3 & ok / L2 (MCP) & \$0.0120 \\
\midrule
\multicolumn{8}{r}{Direct metered cost} & \$0.1912 \\
\multicolumn{8}{r}{Shared allocation (vector store + provisioned pool idle, metered share)} & \$0.0210 \\
\multicolumn{8}{r}{\textbf{Fully loaded} $\hat{c}(b)$ \textbf{(Eq.~\ref{eq:cost})}} & \textbf{\$0.2122} \\
\bottomrule
\end{tabular}
\end{table*}

From the same rows: $\mathrm{RW}(b) = 0.0234/0.2122 \approx 11\%$ (the failed Fusion attempt), $\mathrm{CO}(b) = 0.0066/0.2122 \approx 3\%$, and $\mathrm{TE}(b) = 7{,}900 / 57{,}800 \approx 0.14$ --- output tokens of all successful steps (the failed attempt excluded; this transaction type's versioned definition of ``final'') over all tokens consumed --- flagging that most tokens bought context and coordination, not the answer. Across a synthetic month of O2C transactions of this shape, a close-out report would read: median cost per order \$0.19, $c_{95} = \$0.61$ (long tail driven by exception cases that loop in ServiceNow), rework 9\% of O2C AI spend. At close, reconciliation (Eq.~\ref{eq:recon}) against the settlement ledger shows $V = 7.4\%$ for one provider, decomposed as 1.1\% rate drift, 0.8\% timing, 1.5\% pool idle --- and 4.0\% unmetered: a team calling a model vendor directly on a corporate card, visible in the FOCUS SaaS rows and absent from attribution. That 4.0\% is the shadow-AI finding; it is produced by the ledger delta, not by an audit.

The scenario shows the three answers the business asked for: a cost for \emph{this} transaction, with provenance; the cost stated per order and per process, not per SKU; and a monthly tie-out to the invoice with a named residual.

\section{Analytical Evaluation}
\label{sec:eval}

Absent a field deployment, we evaluate the model analytically: coverage of R1--R7 against the alternatives an enterprise would actually consider (Table~\ref{tab:compare}).

\begin{table*}[!t]
\caption{Requirements coverage by approach (\checkmark{} met, $\sim$ partial, $\times$ unmet).}
\label{tab:compare}
\centering
\begin{tabular}{@{}lccccc@{}}
\toprule
 & \textbf{Tags on} & \textbf{FOCUS} & \textbf{Observ-} & \textbf{Gateway} & \textbf{Two-} \\
 & \textbf{billing} & \textbf{1.4+1.5} & \textbf{ability only} & \textbf{only} & \textbf{Ledger} \\
\midrule
R1 neutrality & $\sim$ & \checkmark & $\sim$ & $\sim$ & \checkmark \\
R2 txn grain & $\times$ & $\times$ & $\sim$ & $\sim$ & \checkmark \\
R3 biz language & $\times$ & $\times$ & $\times$ & $\sim$ & \checkmark \\
R4 agentic completeness & $\times$ & $\sim$ & \checkmark & $\sim$ & \checkmark \\
R5 economic fidelity & $\sim$ & \checkmark & $\times$ & $\sim$ & \checkmark \\
R6 reconcilability & \checkmark & \checkmark & $\times$ & $\times$ & \checkmark \\
R7 timeliness & $\times$ & $\times$ & \checkmark & \checkmark & \checkmark \\
\bottomrule
\end{tabular}
\end{table*}

\textbf{Tags on billing data} inherit settlement grain: they segment workloads and teams, never transaction instances, and cannot vary per request on token-metered SKUs (R2, R3, R7 unmet). \textbf{FOCUS alone, including the announced 1.5 scope}, achieves neutrality, economic fidelity, and reconcilability by design, and 1.5's requester columns add identity grain \cite{tokeneconomics15}; but identity is not transaction (Section~\ref{sec:problem}), billing lags, and process language is absent (R2, R3, R7). \textbf{Observability alone} captures agentic structure and is timely, but carries no settlement truth: no negotiated rates, no amortization, no invoice tie-out --- numbers finance cannot certify (R5, R6). \textbf{Gateway alone} meters what passes through it and prices it in near-real time, but misses what bypasses it, cannot see the invoice, and stops at workload grain unless it also solves correlation --- at which point it has become the attribution ledger of this paper. \textbf{The Two-Ledger Model} meets all seven because it does not ask either dataset to do the other's job: settlement truth stays in FOCUS; transaction truth stays in telemetry; the join, the rate card, and the reconciliation are the contributions.

\textbf{Threats to validity.} This is an architectural argument supported by a simulation on synthetic data (Section~\ref{sec:sim}), not a field study; absolute cost figures in Sections~\ref{sec:scenario} and \ref{sec:sim} are synthetic by design, and the simulation's parameters are chosen rather than measured. Correlation quality at L3/L4 is workload-dependent and unbenchmarked here; the confidence-provenance discipline bounds, but does not eliminate, misattribution. The rate card approximates settlement between closes; Eq.~\ref{eq:recon} measures exactly how well. Finally, two standards the model depends on are still in flux: FOCUS 1.5 is pre-ratification \cite{aispendops15}, and the OpenTelemetry GenAI conventions carry Development status \cite{otel-genai}; the model depends on their column and attribute \emph{roles}, which are stable in direction, more than on their final names.

\section{Simulation Study}
\label{sec:sim}

\emph{All data in this section is synthetic and generated by the study's own code; every usage record carries its true business transaction, so the measurements below are made against known ground truth. The study evaluates the mechanisms relative to one another under controlled conditions. It makes no claim about absolute cost levels in any production estate.}

\subsection{Setup}
A generator produces 20,000 business transactions of three types (order-to-cash, procure-to-pay, claim-to-settlement) across seven systems of record, yielding 109,792 model invocations priced over four SKUs from three providers at the settled list rates of Table~\ref{tab:o2c}. Steps fail and retry with per-step probabilities of 1--12\%; 5\% of transactions are \emph{runaway}, repeating one step a geometrically distributed number of times (mean nine). Instrumentation is controlled by $p_{\mathrm{trace}}$ (share of invocations carrying trace context), $p_{\mathrm{proto}}=0.5$ (share of the remainder carrying A2A/MCP metadata), $p_{\mathrm{event}}$ (availability of business-event feeds for L3), a 5\% key-collision rate, and a confidence floor below which L4 assignments are pooled. L4 scores a weighted similarity over temporal proximity with decay, tenant match, agent-stage plausibility, and lifetime containment; its score-to-confidence mapping is calibrated on a separate 4,000-transaction sample with known truth, in quantile bins with monotone precision, as Section~\ref{sec:spine} prescribes. The random seed is fixed. The code, the results, and every synthetic dataset are available \cite{repo}, together with a script that rebuilds the reconciliation results from the data files alone.

\subsection{E1: Correlation Under Degraded Instrumentation}
\begin{figure}[!t]
\centering
\includegraphics[width=0.95\columnwidth]{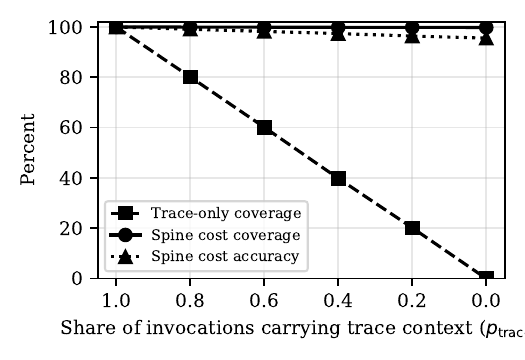}
\caption{E1 --- cost coverage and accuracy as trace propagation degrades (floor 0.80, $p_{\mathrm{event}}=0.85$).}
\label{fig:e1}
\end{figure}

Table~\ref{tab:e1} and Fig.~\ref{fig:e1} compare the spine with a trace-only baseline (what observability alone attributes) as $p_{\mathrm{trace}}$ falls from 1.0 to 0.0. The baseline's coverage falls with it; the spine's cost-weighted coverage stays above 99\%. What degrades is accuracy: cost-weighted accuracy falls from 100\% to 95.6\% as assignment shifts to L3 and L4, and misattributed spend rises to 4.44\% of the total. The second sweep shows what holds accuracy up: with trace context at 40\%, removing the business-event feeds ($p_{\mathrm{event}}=0$) pushes misattribution to 14.21\%, because L3 is the only deterministic mechanism that survives system-of-record boundaries. The third sweep shows the confidence floor as a dial between coverage and certifiability: raising it from 0.80 to 0.95 at $p_{\mathrm{trace}}=0.4$, $p_{\mathrm{event}}=0.5$ moves 15.6\% of spend into explicit pools and takes misattribution from 7.57\% to zero. That is the explicit-degradation discipline of Section~\ref{sec:spine} in numbers: a finance organization can choose a provable 100\% accuracy on 84.4\% of spend, with the rest pooled and labeled, rather than an unprovable 99.6\%.

\begin{table*}[!t]
\caption{E1 --- correlation coverage and accuracy (cost-weighted) versus instrumentation. Trace-only coverage is the share attributable by trace context alone. Pooled: spend assigned to explicit pools. Misattrib.: spend assigned to the wrong transaction. $p_{\mathrm{proto}}=0.5$, 5\% key collisions throughout.}
\label{tab:e1}
\centering\footnotesize
\begin{tabular}{@{}ccccccccc@{}}
\toprule
$p_{\mathrm{trace}}$ & $p_{\mathrm{event}}$ & Floor & Trace-only cov. & Spine cost cov. & Cost acc. & Pooled & L4 share & Misattrib. \\
\midrule
1.0 & 0.85 & 0.80 & 100.0\% & 100.0\% & 100.0\% & 0.0\% & 0.0\% & 0.00\% \\
0.8 & 0.85 & 0.80 & 80.1\% & 100.0\% & 99.1\% & 0.0\% & 2.6\% & 0.89\% \\
0.6 & 0.85 & 0.80 & 60.2\% & 99.9\% & 98.2\% & 0.1\% & 5.2\% & 1.76\% \\
0.4 & 0.85 & 0.80 & 39.8\% & 99.9\% & 97.4\% & 0.1\% & 7.8\% & 2.62\% \\
0.2 & 0.85 & 0.80 & 20.2\% & 99.8\% & 96.4\% & 0.2\% & 10.6\% & 3.59\% \\
0.0 & 0.85 & 0.80 & 0.0\% & 99.8\% & 95.6\% & 0.2\% & 13.1\% & 4.44\% \\
0.4 & 1.00 & 0.80 & 40.0\% & 100.0\% & 99.3\% & 0.0\% & 4.1\% & 0.69\% \\
0.4 & 0.50 & 0.80 & 40.1\% & 99.6\% & 92.4\% & 0.4\% & 16.9\% & 7.57\% \\
0.4 & 0.00 & 0.80 & 40.2\% & 99.3\% & 85.7\% & 0.7\% & 29.4\% & 14.21\% \\
0.4 & 0.50 & 0.85 & 40.0\% & 98.7\% & 93.0\% & 1.3\% & 15.8\% & 6.96\% \\
0.4 & 0.50 & 0.90 & 40.0\% & 97.7\% & 93.4\% & 2.3\% & 15.1\% & 6.47\% \\
0.4 & 0.50 & 0.95 & 39.8\% & 84.4\% & 100.0\% & 15.6\% & 0.0\% & 0.00\% \\
\bottomrule
\end{tabular}
\end{table*}

\subsection{E2: Reconciliation Recovers Planted Effects From Observable Data}
\begin{figure}[!t]
\centering
\includegraphics[width=0.95\columnwidth]{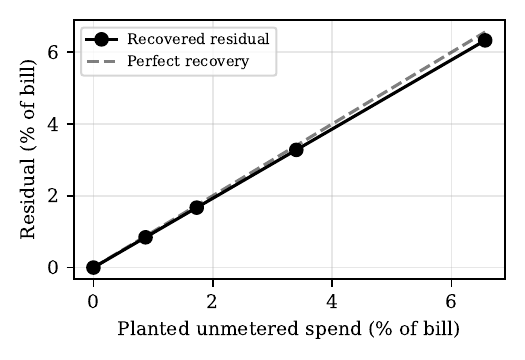}
\caption{E2 --- planted unmetered spend versus the residual the engine recovers from observable data.}
\label{fig:e2}
\end{figure}

The reconciliation engine sees only what it would see in practice: the attribution ledger (metered records with quantities and rate-card estimates), a FOCUS-style billing export with per-SKU, per-day billed quantities and cost, the settled price sheet, and the commitment contract for a provisioned pool. Into the bill the study plants: unmetered consumption of 0--8\% of metered usage (shadow AI), boundary-timing shifts (half of the usage in the last six hours billed in the next period), provider rounding of quantities, a 15\%-idle provisioned-throughput commitment, a 3\% price overbilling on one SKU, and a 2\% omission of usage on another. Table~\ref{tab:e2} and Fig.~\ref{fig:e2} report the decomposition of Section~\ref{sec:recon}. The unmetered residual recovers the planted shadow spend within 0.24 percentage points of the bill across all levels, after rate drift (restatement at the price sheet), timing (last-day quantity reappearing on day one of the next period), and idle (from the contract) are removed. The two-sided check flags the overbilled SKU at exactly the planted amount with no false flags on other SKUs, and the omitted usage appears as a negative residual on its SKU --- the provider billing-discrepancy register of Section~\ref{sec:ext}. The small under-recovery at higher shadow levels comes from provider rounding and from shadow usage that itself sits in the boundary window; both are visible in the decomposition rather than hidden.

\begin{table*}[!t]
\caption{E2 --- reconciliation on observable data. Recovered residual versus planted unmetered spend (both as \% of the bill); decomposition components; two-sided detection of a planted 3\% price overbilling and a 2\% usage omission (USD, one synthetic month).}
\label{tab:e2}
\centering\footnotesize
\begin{tabular}{@{}ccccccc@{}}
\toprule
Planted unmetered & Recovered residual & Error (pp) & Rate drift & Timing & Overbilling flagged / planted & Under-billing residual / planted \\
\midrule
0.00\% & -0.00\% & 0.00 & -1.19\% & 0.40\% & \$16.89 / \$16.89 & \$-2.95 / \$-2.94 \\
0.87\% & 0.84\% & 0.03 & +1.28\% & 0.38\% & \$17.06 / \$17.06 & \$-2.98 / \$-2.97 \\
1.73\% & 1.67\% & 0.06 & +0.61\% & 0.36\% & \$17.23 / \$17.23 & \$-3.01 / \$-3.00 \\
3.40\% & 3.28\% & 0.12 & -0.23\% & 0.30\% & \$17.57 / \$17.57 & \$-3.18 / \$-3.06 \\
6.56\% & 6.33\% & 0.23 & +0.43\% & 0.17\% & \$18.24 / \$18.24 & \$-3.50 / \$-3.18 \\
\bottomrule
\end{tabular}
\end{table*}

\subsection{E3: Mid-Flight Gating Against Runaway Transactions}
\begin{figure}[!t]
\centering
\includegraphics[width=0.95\columnwidth]{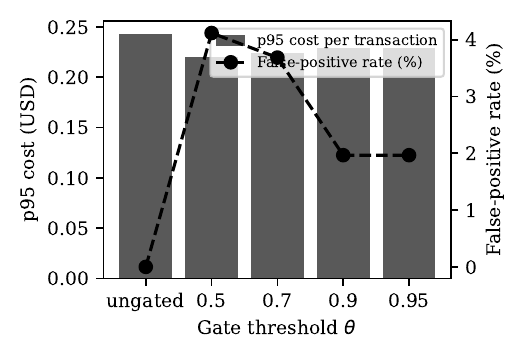}
\caption{E3 --- p95 cost per transaction and false-pause rate by gate threshold.}
\label{fig:e3}
\end{figure}

The forecaster of Section~\ref{sec:ext} is fitted on 12,000 separate training trajectories, with the cost objective per type set at the 95th percentile of non-runaway training cost. Table~\ref{tab:e3} and Fig.~\ref{fig:e3} report, for gate thresholds $\theta$, the cost distribution of the 20,000 test transactions when a transaction is paused after the call at which the forecast breach probability exceeds $\theta$. Ungated, p99 cost per transaction is \$0.701 against a p95 of \$0.243 --- the runaway tail. At $\theta=0.5$ the gate cuts p99 by 61\% and total spend by 9.3\%, catching 74\% of runaways, at the price of pausing 4.1\% of transactions that would have finished within budget; at $\theta=0.9$ the cut is 49\% with a 2.0\% false-pause rate. Thresholds 0.9 and 0.95 give identical results because the forecaster's bin-wise estimates jump from 0.81 to 1.0 at the top accumulated-spend bin, so no gating decision changes between them; a finer forecaster would separate them. The p95 moves little, as expected: it is set by normal transactions, which the gate leaves alone. The threshold is therefore a policy choice between tail protection and approval friction, and the study reports both sides of it.

\begin{table*}[!t]
\caption{E3 --- mid-flight gating. Cost per transaction (USD) and gating outcomes by threshold $\theta$; 5\% of test transactions are runaway.}
\label{tab:e3}
\centering\footnotesize
\begin{tabular}{@{}ccccccc@{}}
\toprule
Gate threshold $\theta$ & p95 cost & p99 cost & Mean cost & Spend saved & False-positive rate & Runaways caught \\
\midrule
ungated & \$0.243 & \$0.701 & \$0.1581 & 0.0\% & 0.00\% & 0.0\% \\
0.5 & \$0.220 & \$0.276 & \$0.1435 & 9.3\% & 4.12\% & 74.4\% \\
0.7 & \$0.224 & \$0.307 & \$0.1467 & 7.2\% & 3.69\% & 74.4\% \\
0.9 & \$0.229 & \$0.358 & \$0.1484 & 6.2\% & 1.97\% & 72.0\% \\
0.95 & \$0.229 & \$0.358 & \$0.1484 & 6.2\% & 1.97\% & 72.0\% \\
\bottomrule
\end{tabular}
\end{table*}

\subsection{What the Study Shows and Does Not Show}
It shows that the mechanisms behave as the architecture claims under controlled conditions: the spine trades accuracy, not coverage, as instrumentation degrades, and the confidence floor converts that trade into an explicit, labeled pool; the residual is recoverable from data a FinOps team already holds, in both directions; and forecast-based gating bounds the tail at a measurable approval cost. It does not show production magnitudes. The generator's parameters --- collision rates, feed availability, runaway frequency, provider rounding --- are chosen, not measured, and a field pilot remains the next step (Section~\ref{sec:discussion}).

\section{Realization: Product, Assembly, and a Best-of-Breed Exemplar}
\label{sec:realization}

The architecture of Section~\ref{sec:tlm} raises the question every adopting enterprise and every product strategist asks next: \emph{must someone build this as a product, or do customers assemble it from what they already own?} The answer is that the contracts make both possible; that is what designing to FOCUS, OpenTelemetry GenAI, and W3C Trace Context buys. The economics then decide where a product earns its place. This section examines three sourcing models, then grounds them in an illustrative best-of-breed mapping of the eight components onto the October 2026 market.

\subsection{Three Sourcing Models}
Table~\ref{tab:sourcing} compares the realizations. \emph{Self-assembly} has the enterprise platform team build the join on its own lakehouse; it suits organizations with mature FinOps and data platforms, and its hidden cost is permanent: someone must maintain the rate card and the correlation engines forever. A \emph{delivered accelerator} has an integrator implement the reference architecture as code the enterprise then owns; it buys the architecture without a new vendor, at project cost and with ownership transferring at handover. A \emph{productized join layer} ships components 5 and 7 --- the only two the market does not sell (Section~\ref{sec:eval}) --- as software deployed in the customer's account, writing the attribution ledger to the customer's lakehouse, read-only to billing, with every adjacent component bring-your-own. Its adoption logic for customers with existing investments is simple: nothing already owned is replaced, and the first deliverable (the reconciliation and shadow-AI report) requires only data the customer already has. Because all three models exchange the same records in the same open formats, an enterprise can start in one and move to another; the deciding variables are who carries the rate-card maintenance burden and who certifies correlation confidence to finance.

\begin{table*}[!t]
\caption{Sourcing models for realizing the Two-Ledger Model.}
\label{tab:sourcing}
\centering\footnotesize
\setlength{\tabcolsep}{4pt}
\begin{tabular}{@{}p{2.5cm}p{4.3cm}p{4.3cm}p{4.3cm}@{}}
\toprule
 & \textbf{Self-assembly} & \textbf{Delivered accelerator} & \textbf{Productized join layer} \\
\midrule
Who builds components 5 and 7 & Enterprise platform team, on the existing lakehouse & Integrator, from a reference implementation; enterprise owns it after handover & Vendor product, deployed in the customer's account and lakehouse \\
Time to first value & Quarters & About a quarter & Weeks: reconciliation and shadow-AI report from existing billing and gateway data, before any instrumentation \\
Cost shape & Internal engineering plus a permanent rate-card and engine maintenance burden & Project cost; maintenance transfers to the enterprise & Subscription, priced on attributed spend or metered transactions; vendor maintains the rate card \\
Lock-in surface & None, but schema-drift risk against evolving standards & Low; open formats throughout & Bounded by the contracts: FOCUS and OTel GenAI in, FOCUS-shaped rows and gate decisions out \\
Best fit & Large platform organizations with FinOps and data maturity & Enterprises wanting the architecture without a new vendor relationship & Multi-cloud, multi-gateway estates needing finance-grade tie-out and a maintained price book \\
\bottomrule
\end{tabular}
\end{table*}

\subsection{A Best-of-Breed Exemplar}
Table~\ref{tab:exemplar} maps each component onto illustrative market realizations as of October 2026, drawn from practitioner tool surveys and vendor documentation \cite{roundup-sp,roundup-holori,futurum-omni,msoft-fabric}. The mapping is point-in-time and illustrative, not an evaluation or endorsement; two of the surveyed sources are vendor-published, and the table should be re-verified at reading time. The analytical point stands despite those caveats: seven of the eight components are available as products or commodity infrastructure, assembled across at least four distinct vendor categories, while component 5 has no standalone vendor and component 7 has none at all --- which is exactly where Table~\ref{tab:compare}'s coverage gaps lie.

\begin{table*}[!t]
\caption{Illustrative best-of-breed realization of the eight components (market snapshot, October 2026; examples, not endorsements) \cite{roundup-sp,roundup-holori,futurum-omni,msoft-fabric}.}
\label{tab:exemplar}
\centering\footnotesize
\setlength{\tabcolsep}{4pt}
\begin{tabular}{@{}p{2.4cm}p{5.4cm}p{4.2cm}p{3.9cm}@{}}
\toprule
\textbf{Component} & \textbf{Exemplar realizations} & \textbf{What they cover} & \textbf{What remains outside} \\
\midrule
1\; LLM gateway & MuleSoft Omni Gateway; Kong AI Gateway; LiteLLM; Portkey; provider-native endpoints & Token capture and limits, guardrails, MCP/A2A policy, identity propagation, correlation identifiers & Prices, invoices, and anything that bypasses the proxy \\
2\; Backfill emitters \& pollers & OpenTelemetry GenAI SDKs and Collector; provider usage and billing APIs & Paths no proxy sees (embedded copilots, provider-managed agents, batch) & Lower confidence than gateway metering, by construction \\
3\; Attribution ledger & Iceberg/Delta tables on Databricks, Snowflake, or BigQuery & Durable, queryable invocation-grain store & The FOCUS-shaped \texttt{x\_} profile is a schema discipline, not a product \\
4\; Settlement lake & Apptio, CloudZero, Vantage, Finout; Microsoft FinOps toolkit & FOCUS ingestion, invoices, reconciliation to cash & Runtime usage and the transaction boundary \\
5\; Rate card service & --- (build; gateway price maps seed it) & --- & Negotiated rates, commitment amortization, and tier logic have no standalone vendor; the FOCUS 1.5 price sheet will anchor, not replace, it \\
6\; Taxonomy \& identity registry & Entra ID / Okta / SPIFFE; Purview / Collibra / DataHub; APQC PCF content \cite{apqc} & Workload and agent identity, catalogs, process codes & One key schema across traces, tags, and gateway metadata is governance work, not a purchase \\
7\; Correlation, allocation, reconciliation & --- (the white space); partial pieces: unit-cost instrumentation, virtual tagging, invoice-ingesting observability & Fragments on one side of the grain gap or the other & The trace-to-invoice join at transaction grain, with confidence provenance and a decomposed $\Delta$ \\
8\; Unit economics \& gates & Unit-cost analytics; BI semantic layers; OPA-class policy; gateway budget enforcement & Analytics and enforcement actuators & The closed loop from $c_{95}$ breach to gate decision \\
\bottomrule
\end{tabular}
\end{table*}

\subsection{The Strongest Edge Still Stops at the Ledger Boundary}
The pattern is sharpest in the most complete metering stacks. MuleSoft's Omni Gateway (the former Flex Gateway), with Agent Fabric, spans LLM routing and token limits, prompt guardrails and PII detection, OpenAPI-to-MCP conversion, end-to-end correlation identifiers, consumption telemetry, and an agent registry, broker, and visualizer with A2A and MCP support \cite{futurum-omni,msoft-fabric}. In this paper's terms that is a strong component 1, parts of 2 and 6, and the enforcement actuator of 8 --- and, because the surrounding integration platform already connects the systems of record, a natural emitter of the spine's L3 business events. Yet its public materials describe FinOps only as a consumer of its telemetry: no invoice ingestion, no rate card, no FOCUS dataset, no reconciliation \cite{futurum-omni}. The most capable edge on the market still stops at the ledger boundary. That boundary is not a vendor oversight. It is the grain mismatch of Section~\ref{sec:problem}, and crossing it is the job of components 5 and 7 --- the productizable core.

\section{Discussion}
\label{sec:discussion}

\textbf{Governance integration.} Transaction-grain cost turns AI governance from policy into arithmetic. Budget gates become executable: an agent, workflow, or process carries a cost SLO ($c_{95}$ per outcome type), enforced at the gateway, with breaches routed through the same lifecycle gates that govern capability and risk --- the Govern and Gate stages of an agentic governance maturity progression \cite{aagmm}. Coordination overhead (Eq.~\ref{eq:co}) prices multi-agent deliberation itself, making consensus mechanisms a cost-benefit decision rather than an architectural default \cite{scf}. The reconciliation residual makes shadow-AI detection a monthly close activity, owned jointly by finance and the AI platform team.

\textbf{Agent identity and settlement-native attribution.} FOCUS 1.5's requester columns call for a stable, enterprise-wide agent identity scheme used identically in both ledgers --- the same \texttt{PrincipalId} in the gateway record and, where providers support it, in the billing row. Looking further out, verifiable agent identity with agent-held credentials and payment instruments \cite{patent} points to a settlement-native future in which the agent's own transaction record is the attribution row; the Two-Ledger join is the bridge architecture until then.

\textbf{Standardization outlook.} The attribution ledger published as FOCUS-shaped, invocation-grain rows with \texttt{x\_} business keys (Section~\ref{sec:tlm}) is deliberately also a proposal: it shows, in the specification's own extension mechanism, what an \emph{invocation-grain usage profile} of FOCUS would look like. We intend to contribute the mapping of Table~\ref{tab:schema} to the FOCUS and OpenTelemetry GenAI working groups as alignment input, so that the columns finance trusts and the attributes engineering emits converge on one vocabulary for AI consumption.

\textbf{Limitations and future work.} The immediate next step is a field pilot measuring attribution coverage, correlation precision at each spine level, and reconciliation variance over several closes --- the empirical validation this paper does not claim. Further work includes correlation benchmarks for L4 matching, treatment of fine-tuning and inference-adjacent costs (embedding refresh, evaluation runs) as transaction-amortizable overhead, and re-validation of the FOCUS 1.5 and OpenTelemetry mappings at ratification.

\section{Conclusion}
\label{sec:conclusion}

Businesses account for work in transactions and outcomes. AI providers bill in SKUs and periods. In three years, FOCUS has given the bill one shape across clouds, SaaS, and AI vendors, and its forthcoming AI scope will name models, tokens, and even agents inside billing data. One thing no billing standard can contain is the enterprise's own transaction boundary --- the order, the claim, the case --- because the provider never sees it. We have argued that this makes transaction-grain AI cost a \emph{sub-ledger} problem. Keep settlement truth in a FOCUS ledger. Capture transaction truth in an invocation-grain attribution ledger at the gateway. Correlate across systems of record through a layered spine with explicit provenance. Price through a versioned rate card. Report the result in the process language the business already uses, and reconcile the two ledgers every close, letting the residual name the shadow AI. Under simulation with known ground truth, the join keeps cost coverage above 99\% where tracing alone collapses, recovers planted unmetered spend within 0.24 percentage points, and bounds the runaway tail at a measurable approval cost. The bill says what the enterprise spent. The trace says why. The join answers the question the business asked.

\section*{Acknowledgment and AI-Use Disclosure}
Generative AI assistance (Anthropic Claude) was used for literature synthesis, drafting, simulation and figure code, and LaTeX preparation under the author's direction. The author reviewed, edited, and takes full responsibility for all content; all references were verified by the author against their primary sources prior to submission.

\end{document}